\def\year{2022}\relax
\documentclass[letterpaper]{article} % DO NOT CHANGE THIS
\usepackage{aaai22}  % DO NOT CHANGE THIS
\usepackage{times}  % DO NOT CHANGE THIS
\usepackage{helvet}  % DO NOT CHANGE THIS
\usepackage{courier}  % DO NOT CHANGE THIS
\usepackage[hyphens]{url}  % DO NOT CHANGE THIS
\usepackage{graphicx} % DO NOT CHANGE THIS
\usepackage{natbib}  % DO NOT CHANGE THIS AND DO NOT ADD ANY OPTIONS TO IT
\usepackage{caption} % DO NOT CHANGE THIS AND DO NOT ADD ANY OPTIONS TO IT
\DeclareCaptionStyle{ruled}{labelfont=normalfont,labelsep=colon,strut=off} % DO NOT CHANGE THIS
\usepackage{algorithm}
\usepackage{algorithmic}
\usepackage{wrapfig}

\usepackage{newfloat}
\usepackage{listings}
\floatstyle{ruled}
\newfloat{listing}{tb}{lst}{}
\floatname{listing}{Listing}
 
\usepackage{multirow}

\title{Polarization and Morality: 
Lexical Analysis of Abortion Discourse on Reddit}
\author {
    Tessa Stanier,
    Hagyeong Shin
}
\affiliations {
   Department of Linguistics\\
    University of California San Diego\\
    \texttt{testanie@ucsd.edu}, \texttt{hashin@ucsd.edu}
}

\nocopyright 

\usepackage{bibentry}

\begin{document}

\maketitle

\begin{abstract}
This study investigates whether division on political topics is mapped with the distinctive patterns of language use. 
We collect a total 145,832 Reddit comments on the abortion debate and explore the languages of subreddit communities \textit{r/prolife} and \textit{r/prochoice}. With consideration of the Moral Foundations Theory, we examine lexical patterns in three ways. 
First, we compute proportional frequencies of lexical items from the Moral Foundations Dictionary in order to make inferences about each group's moral considerations when forming arguments for and against abortion. 
We then create n-gram models to reveal frequent collocations from each stance group and better understand how commonly used words are patterned in their linguistic context and in relation to morality values.
Finally, we use Latent Dirichlet Allocation to identify underlying topical structures in the corpus data.
Results show that the use of morality words 
is mapped with the stances on abortion.
\end{abstract}

\section{Introduction}
\noindent 
Firth's quote ``You shall know a word by the company it keeps'' \cite{firth1957synopsis} is often cited to suggest that the context of the word can tell us about the word itself; it has been studied that the meaning of the word can be assessed by the co-occurring words \cite{landauer1998introduction}.
So could the distribution of words also reflect the real-world division between speakers?
Linguistic patterns have been suggested as a reflection of speaker's emotional state \cite{pennebaker2003psychological} and political identity \cite{ashokkumar2022tracking}. Based on these findings, the current study investigates whether a division on a political topic could be reflected by patterns of words used by two groups with distinctive stances on the topic.

Abortion rights as outlined by Roe v. Wade (1973) and the debate around this case showcase the significance of abortion in American politics.
There was a resurgence in public discussion following the U.S. Supreme Court draft opinion leak in May of 2022 and subsequent decision to overturn Roe v. Wade in June 2022. 
This heightened public awareness has brought the discussion to new salience; the event drew substantial attention, as abortion is a deeply personal topic and is often considered as an issue of morality vs immorality. With the neat division between pro-choice liberals and pro-life conservatives, and the divisive nature of the topic, abortion rhetoric serves as an optimal case to examine the polarization in political language.

We approach the conversation around abortion with the goal of defining the lexical patterns of two opposing groups, namely the pro-choice and pro-life. The two groups' conversation is collected from Reddit.
We focus on how the underlying ideas on morality are reflected through each group's patterns of lexical usage, using the 
Morality Foundation Dictionary \cite{graham_liberals_2009-1} as a tool for analysis.
We first compare the frequency of the morality words appearing in each group's conversation to see how the two groups lean into different morality values. After that, we move onto n-gram models to look into the context of the words. Results from n-gram models reveals the co-occurrence of the words, thus showing what kind of arguments that each group constructed with the morality words.
Lastly, we perform topic modeling. 
Results showed that two groups used morality words to the same ratio to stay with the topic under discussion, but they used the morality words in different context and topics to form opposing arguments and appeal to different morality values.

\section{Background}

\subsection{Language as an Expression of Political Identity}

Language use can vary drastically across social groups \cite{eckert_limits_2019}, including groups with different political orientations \cite{ashokkumar2022tracking}.
People with different stances on an issue may present different patterns of language that are representative of that stance \cite{trott_reconstruing_2020}. For example, \citet{greene-resnik-2009-words} found that the choice between active and passive form is relevant in determining speaker's sentiment about violent events. The phrasing \textit{she murdered the guard} as opposed to \textit{the guard was murdered}, places \textit{she} as the subject, emphasizing the role played in the murder, whereas the latter sentence places focus on \textit{the guard}. On the lexical level, this can be observed with the choice of the term \textit{illegal alien} as opposed to \textit{illegal immigrant}---while these terms may be used interchangeably without any lapse in communication, the former phrasing effectively de-emphasizes the humanity of the referent, which could be representative of the speaker's views on immigration.

\subsection{Linguistic Features of Morality}

While some may assume that people develop their stances on political and social issues after careful deliberation and thought, it is more common for us to experience immediate emotional reactions and then subsequently search for arguments that support them \cite{haidt_righteous_2012}. 
\citet{haidt_when_2007} created the Moral Foundations Theory (MFT) to explain stark ideological differences between liberals and conservatives in terms of morality. Authors identify five moral foundations, shown in Table \ref{tab:mfd}.

\begin{table}[t]
\footnotesize
    \centering
    \begin{tabular}{|l|l|}
    \hline
        MFD category & Explanation and example words \\ \hline
        \multirow{2}{*}{Harm/Care}  & Care for others and aversion to pain \\
        & \textit{safe, peace, care, harm, suffer, war}\\\hline 
        \multirow{2}{*}{Fairness/Cheating}
         & Fairness, reciprocity and justice  \\
         & \textit{fair, equal, honest, unfair, bias, unjust} \\ \hline
        \multirow{2}{*}{Loyalty/Betrayal} 
         & Loyalty to one's in-group \\
         & \textit{family, nation, loyal, foreign, treason, spy} \\ \hline
        \multirow{2}{*}{Authority/Subversion}
         & respect for social hierarchies  \\
         & \textit{obey, duty, law, rebel, dissent, defy} \\ \hline
         \multirow{2}{*}{Sanctity/Degradation}
         & aversion to disease and impurity\footnote{While only five foundations have thus far been endorsed, there is preliminary evidence for a sixth foundation: Liberty/Oppression. We have not included this candidate in our analysis, and there is no such category available in the MFD at the time of this study.}  \\
         & \textit{pure, clean, virtuous, disgust, sin, defile} \\ 
         \hline
    \end{tabular}
    \caption{Five MFD categories considered in our study. Explanations and example words for each categories are from \citet{graham_liberals_2009-1}.}
    \label{tab:mfd}
\end{table}

\citet{graham_liberals_2009-1} have found that people who identify as liberal tend to hold the moral foundations of \textit{harm/care} and \textit{fairness/reciprocity} of heightened consideration, while those who identify as conservative tend to consider all five foundations more or less equally. \citet{koleva_tracing_2012} further elaborated on how endorsement of the various moral foundations can accurately predict individuals positions on controversial topics in modern culture wars; the authors demonstrate that moral intuitions as defined by the MFT accurately predict individuals’ stance on topics such as abortion, gun control and capital punishment.

More recently, scholars have been expanding prior methods for measuring moral intuitions by incorporating computational models and applying them to large corpora \cite{sagi_measuring_2014, fulgoni_empirical_2016}. \citet{graham_liberals_2009-1} proposed the Moral Foundations Dictionary (MFD), a collection of words and word stems that are related to the various moral foundations, which has been used in combination with the Linguistic Inquiry and Word Count \cite{pennebaker_development_2015} to facilitate analysis of large collections of representative text data. \citet{sagi_identifying_2013} utilized NLP methods that calculate the semantic similarity of MFD items and measure their frequency in republican vs. democrat speeches on abortion, finding that while democrats are most concerned with issues of fairness, republicans primarily referenced purity.

\section{Morality Words in the Abortion Debate}
\subsection{Research Question and Hypotheses}
In the present work, we explore ideological polarization and language usage online by utilizing NLP methods to identify lexical features indexing moral values and topical considerations made by polarized communities, using the abortion debate on Reddit as a case study. With arguments surrounding abortion ranging from fetal person-hood, to adherence to religious dogmas and overt discussion of a mother’s morality, proponents and opponents frequently utilize highly emotional arguments with heavy emphasis on right vs. wrong, making this an ideal topic for analysis. 
We focus on the choices of lexical items as the observable linguistic patterns. 
We expect that groups with different stances will have different ways of using words that are loaded with morality values. 

\subsection{Data}
We've chosen Reddit as a source of data and selected two subreddits with opposing stances, r/prolife and r/prochoice. 
The corpus consists of user comments beginning from May 15th, 2022 until the date of collection on August 7th, 2022, a period when the conversation around abortion was heightened due to the Roe v. Wade overturn. The start date that we have selected is two weeks after the leak of the Supreme Court draft opinion to overturn Roe v Wade, which brought the abortion debate to salience in the public sphere, up until the date of collection. We have only extracted comment data for these purposes and have excluded content from submission posts. In sum, 100,002 comments were retrieved from r/prolife and 45,830 comments from r/prochoice, cummulating in a grand total of 145,832 comments.  All data was taken from the Reddit Pushshift API \cite{baumgartner_pushshift_2020}.

\begin{table*}[t]
\footnotesize
    \centering
    \begin{tabular}{|c|l|l|l|l|}
    \hline
    & \multicolumn{2}{c|}{bigrams} & \multicolumn{2}{c|}{trigrams}\\ \hline
       rank & r/prochoice & r/prolife & r/prochoice & r/prolife \\ \hline
       1 & \textit{birth}, \textit{control} &  \textit{human}, \textit{life} & \textit{roe}, \textit{v}, \textit{wade} & \textit{roe}, \textit{v}, \textit{wade} \\
       
       2  &\textit{year}, \textit{old} &  \textit{birth}, \textit{control} & \textit{10}, \textit{year}, \textit{old} & \textit{life}, \textit{begin}, \textit{conception}  \\
       
       3  & \textit{bodily}, \textit{autonomy} & \textit{bodily}, \textit{autonomy} & \textbf{right}, \textit{bodily}, \textit{autonomy} & \textit{abortion}, \textit{case}, \textit{rape}  \\
       
       4  & \textit{give}, \textit{birth} & \textbf{right}, \textit{life} & \textit{forced}, \textit{give}, \textit{birth} &  \textit{prolifers}, \textit{think}, \textit{abortion}  \\
       
       5  & \textit{human}, \textbf{right} & \textit{human}, \textbf{right} & \textit{website}, \textit{help}, \textit{find} & \textit{think}, \textit{abortion}, \textit{case}  \\
       
       6   & \textit{pregnant, person} & \textit{year}, \textit{old} & \textit{late}, \textit{term}, \textit{abortion} &  \textit{due}, \textit{prolife}, \textit{prolifers} \\
       
       7  & \textit{supreme, court} & \textit{makes, sense} & \textit{site}, \textit{offer}, \textbf{legal} & \textit{prolife}, \textit{prolifers}, \textit{think}  \\
       
       8 & \textit{im, sorry} & \textit{human, being}& \textit{life}, \textit{begin}, \textit{conception} & \textit{10}, \textit{year}, \textit{old}  \\
       
       9 & \textit{woman}, \textbf{right} & \textit{pregnant, woman} & \textit{offer}, \textbf{legal}, \textit{help} &  \textit{foster}, \textbf{care}, \textit{system}  \\
       
       10  & \textit{abortion, right} & \textit{ectopic, pregnancy} & \textbf{legal}, \textit{help}, \textit{anyone}& \textit{right}, \textit{bodily}, \textit{autonomy}  \\
       
       11 & \textit{roe, v} & \textit{unborn, child} & \textit{foster}, \textbf{care}, \textit{system} & \textit{maternal}, \textit{mortality}, \textit{rate} \\
       
       12 &  \textit{forced, birth} & \textit{life, begin}& \textit{crisis}, \textit{pregnancy}, \textit{center} & \textit{late}, \textit{term}, \textit{abortion}  \\
       
       13 & \textit{makes, sense} & \textit{foster}, \textbf{care} & \textit{another}, \textit{person}, \textit{body} & \textit{value}, \textit{human}, \textit{life} \\
       
       14 & \textit{v, wade} & \textit{give, birth} & \textit{state}, \textit{federal}, \textit{government} & \textit{human}, \textit{life}, \textit{begin} \\
       
       15 & \textit{forced, birthers} & \textbf{kill}, \textit{child} & \textit{body}, \textit{without}, \textit{consent} & \textit{bodily}, \textit{autonomy}, \textit{argument}  \\
       
       16 & \textit{10, year} & \textbf{kill}, \textit{baby} & \textit{right}, \textit{people}, \textit{need} & \textit{choice}, \textit{abortion}, \textit{ideology}  \\
       
       17 & \textit{pregnant, woman} & \textit{clump, cell} & \textit{guaranteeing}, \textbf{right}, \textit{abortion} & \textit{rant}, \textit{without}, \textbf{attacked} \\
       
       18  & \textit{people}, \textit{need} & \textit{another}, \textit{human} & \textit{thing, prolifers, image} & \textbf{attacked}, \textit{belief}, \textit{idea}  \\
       
       19  & \textit{ectopic, pregnancy} & \textit{think, abortion} & \textit{sex, marriage, contraception} & \textit{abortion}, \textit{ideology}, \textbf{respect} \\
       
       20  & \textit{life, begin} & \textit{planned, parenthood}  & \textit{amendment}, \textit{guaranteeing}, \textbf{right} & \textit{idea}, \textit{ignorance}, \textit{reason} \\ \hline
    \end{tabular}
    \caption{Top 20 most frequent bigrams and trigrams from r/prolife and r/prochoice. MFD words are boldfaced.}
    \label{tab:ngram}
\end{table*}

\section{Results}
\subsection{Frequency}

We calculated word frequency scores for MFD words, proportional to the total amount of words in each corpus. From this we analyzed scores to determine whether or not the proportional frequencies demonstrated statistically significant differences of MFD words across the two communities that would index different moral concerns. 
Contrary to expectation, no significant differences were observed relating to the various moral concerns as outlined by MFT. While r/prolife had higher frequencies in each of the moral categories with the exception of \textit{loyalty/betrayal}, text data taken from comments in r/prochoice and r/prolife displayed no statistically significant differences in the frequency of words from the MFD. 

However, r/prolife had higher rates of foundation violating words like \textit{murder, harm} and \textit{kill}, which illustrates the common ideological perspective of abortion as murder. Conversely, foundation affirming words like \textit{care, safe, peace} and \textit{protect} were higher in r/prochoice. Aside from this, both subreddits proved to utilize words from each moral category at similar rates.

\subsection{N-grams}

N-gram models revealed more granularity than frequency, because they added context to individual words.
We looked into the top 20 bigrams and trigrams and assessing whether or not the differences in frequent collocations indexes any latent information. The results are presented in the Table \ref{tab:ngram}. 

R/prochoice lexical patterns reflected the focus on women's rights to abortion and inclusion of other affected groups. Their bigrams included (6th: \textit{pregnant}, \textit{person}), (9th: \textit{woman}, \textit{right}), (10th: \textit{abortion}, \textit{right}), (12th: \textit{forced}, \textit{birth}), and (15th: \textit{forced}, \textit{birthers}). Their trigrams included 
(4th: \textit{forced}, \textit{give}, \textit{birth}). (13th: \textit{Another}, \textit{person}, \textit{body}) and (15th: \textit{body}, \textit{without}, \textit{consent}) both reflect the central idea that women should be able to have full autonomy over what happens to their bodies, rather than the government. 
Altogether, these represent the perspective held by pro-choice groups that banning abortion is not actually saving lives, but rather simply forcing women to give birth. 

From r/prolife n-grams, focus shifted towards considerations of life. Their bigrams (15th: \textit{Kill}, \textit{child}) and (16th: \textit{kill}, \textit{baby}) were noteworthy in that they were statements unique to r/prolife rhetoric, emphasizing harm from abortions in a way that elicits a strong emotional reaction from listeners or readers with the word \textit{kill} from the MFD's \textit{care/harm} category. 
Their trigrams reflected the consistent argument, having more terms addressing the value of the fetus' life: (13th: \textit{value}, \textit{human}, \textit{life}) and (14th: \textit{human}, \textit{life}, \textit{begin}).

\begin{figure*}[t]
  \includegraphics[width=1\textwidth]{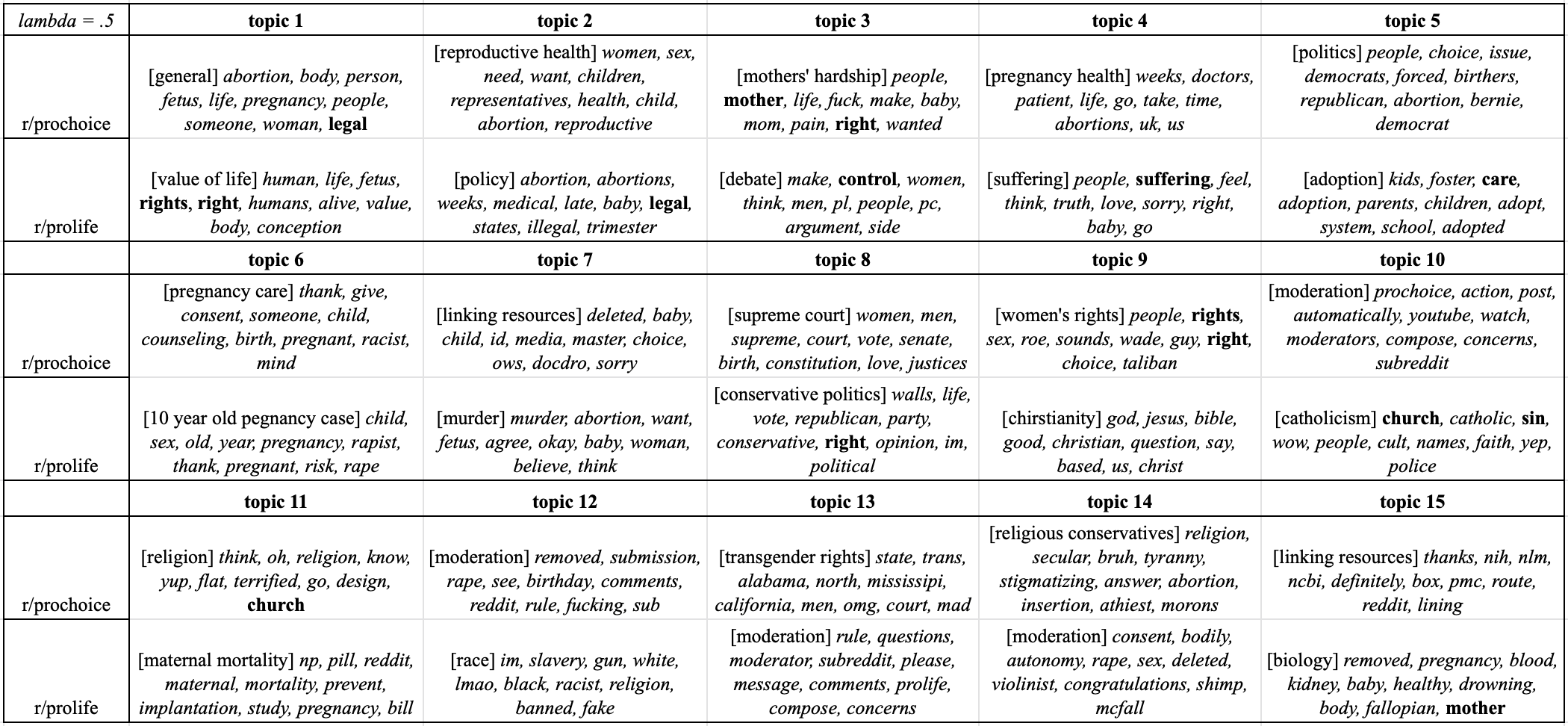}
  \caption*{Table 3: 15 topics extracted from r/prochoice and r/prolife comments. Brackets in each cell indicate a label of a topic selected by researchers. Terms in each topic are italicized and MFD words are boldfaced.}
  \label{fig:lda}
\end{figure*}

A few MFD words arose within n-gram analysis. There were two foundation violating \textit{care/harm} terms in r/prolife: \textit{kill} and \textit{attack}, each of which occurred in two items. \textit{Respect} is also noteworthy in that it belongs to the MFD category of \textit{authority/subversion}. The other \textit{authority/subversion} term that appeared was \textit{legal}, having occurred three times as part of a common phrase in r/prochoice. While \textit{care} did occur in both groups, it was part of a larger phrase, the scope of which we did not consider in our analysis. The last word that appeared was \textit{rights} which falls under the category of \textit{fairness/cheating}. \textit{Rights} is also commonly used in reference to different entities, the mother or the fetus.

\subsection{LDA}

Given the n-gram results, we moved on to look into the further context of the words. We used Latent Dirichlet Allocation (LDA) to uncover topical structures \cite{blei_latent_2003} formed by each group. 
We ran topic models with \textit{k} (the number of the topics) ranging from 3 to 20 with $\lambda$ (weight of a term under a topic) set between .3 and .7 then manually examined which value would result in the least redundancy and highest coherence. In the end, we set \textit{k} to 15 with $\lambda$ set to .5 to run the analysis on the text data from each group. We took into consideration the groups of topic words and granted each \textit{k} a descriptive title that is representative of its content. The results are presented in Table 3.

Discussion from each stance group involved unique combinations of word co-occurrences. R/prochoice hosted topics 1, 5 and 8 which discuss the abortion, politics and the supreme court, with reference to political parties, candidates and actions. Topics 2, 4 and 6 related to women's reproductive healthcare. While topics 12 and 17 contain individual items may not appear meaningful at first glance, further investigation into the top terms of these categories revealed that the strings of letters are pieces of bigger URLs belonging to various websites offering education on reproductive and legislative issues in addition to resources for women affected by the Roe v Wade ruling. There were also topics that expressed concern for the well-being of the various groups that will be affected by the ruling, such as mothers in topic 3, women in topic 9 and transgender individuals in topic 13. In general, topics trend more towards addressing overturning of Roe and its effects on people who can get pregnant. 

Some overall topics were closely connected with the moral foundation of \textit{care/harm}, and this relationship is slightly different for each stance group. When looking at the issue, r/prochoice approached the conversation from a standpoint of providing care to those who no longer have access to abortion after the ruling, as was demonstrated by the various topics discussing healthcare, the rights of groups and providing resources (2, 3, 4, 6, 7, 9, 13, 15). R/prolife conversely hosted several topics that focused on harm resulting from abortions (4, 7, 11). 
The biggest argument of r/prolife group appeared in topics was the value of life, which addresses the issue of fetal personhood, representing the core belief that life begins at conception.
These results indicate that r/prochoice places high importance on providing care, while r/prolife seeks primarily to prevent harm.

There was also greater diversity in topics for r/prolife. While r/prochoice had numerous topics geared towards resources and advocacy, r/prolife additionally hosted topics addressing the group identities. Topics 8, 9, 10 and arguably 12 each isolated specific features that an individual may consider to be apart of the collective group identity. Group morals, such as those outlined by Christianity or Conservatism, may be brought up as justification behind author stance on abortion, or sides may be referenced in order to frame other aspects of the debate. This supports findings from the MFT showing that conservatives score higher in the foundations of \textit{loyalty/betrayal} and \textit{authority/subversion}, which require attention and adherence to one's group membership and the hierarchies within. R/prolife topic 10 contained two \textit{sanctity/degradation} words: \textit{sin} and \textit{church}, aligning with conservative tendency to value this foundation. 
While religious topics were expected to come from r/prolife due to the associated ideology, r/prochoice also discussed religion in topics 11 and 14; although this was likely tied into discussion of the opposing stance group, the MFD word \textit{church} appears in topic 11, as well.

Overall, the LDA results showed different topics arose from the two groups, including different usage of morality words from the MFD dictionary.

\section{Summary}

In this study, we have shown that political polarization is reflected in lexical patterns relative to each stance group.
No differences in frequency of morality words may indicate that speakers must use certain words to stay with the topic under the discussion or to dispute other's arguments. N-gram and LDA results showed further context of the morality words used by two groups. The results showed that each group utilized \textit{care/harm} dimension of morality differently. Certain sets of terms and phrases, such as \textit{forced birth} or \textit{pregnant person} for r/prochoice or \textit{kill child/baby} and \textit{unborn child} for r/prolife, reflect contrasted morality values. 
From such exploration, we see that lexical features and patterns could be used as empirical evidence to illustrate different stances on disputed topics.

\section{Ethics Statement}

Data was collected without any direct interaction with individual users on Reddit and usernames were deleted upon data collection. Final dataset used in the research does not contain any information nor association with individual users.
The paper does not pose any negative impact, and hopefully could contribute to promote bridging the gap in debated issues by increasing awareness of language use.

\bibliography{abortion.bib}

\appendix

\end{document}